\documentclass[12pt]{article}
\usepackage{arxiv}
\usepackage[utf8]{inputenc} % allow utf-8 input
\usepackage[T1]{fontenc}    % use 8-bit T1 fonts
\usepackage{graphicx}
\usepackage{amsmath,amssymb,amsthm}
\usepackage{mathtools}
\usepackage{xspace}
\usepackage{epsfig}
\usepackage[vlined,linesnumbered,ruled,titlenotnumbered]{algorithm2e}
\usepackage{color}
\usepackage{url}
\usepackage{tikz}
\usepackage{float}
\usepackage{booktabs}
\usepackage{makecell}
\usepackage{multicol}
\usepackage{multirow}
\usepackage{longtable}
\usepackage{array}
\usepackage{algorithmic}
\usepackage[libertine]{newtxmath}
\usepackage{colortbl,soul,xcolor,bbm}
\usepackage{subcaption}
\usepackage{nicefrac}
\usepackage{microtype}      % microtypography

\DeclareMathOperator*{\argmax}{\arg\!\max}

\newcommand{\ignore}[1]{}

\begin{document}

\author{Anh Viet Do
\\Optimisation and Logistics\\
School of Computer Science
\\The University of Adelaide
\And
Mingyu Guo
\\Optimisation and Logistics\\
School of Computer Science
\\The University of Adelaide
\And
Aneta Neumann
\\Optimisation and Logistics\\
School of Computer Science
\\The University of Adelaide
\And
Frank Neumann
\\Optimisation and Logistics\\
School of Computer Science
\\The University of Adelaide
}
% Working title
\title{Niching-based Evolutionary Diversity Optimization for the Traveling Salesperson Problem}
\renewcommand{\shorttitle}{Niching-based Evolutionary Diversity Optimization for the TSP}
\date{}
\maketitle
\begin{abstract}
In this work, we consider the problem of finding a set of tours to a traveling salesperson problem (TSP) instance maximizing diversity, while satisfying a given cost constraint. This study aims to investigate the effectiveness of applying niching to maximize diversity rather than simply maintaining it. To this end, we introduce a 2-stage approach where a simple niching memetic algorithm (NMA), derived from a state-of-the-art for multi-solution TSP, is combined with a baseline diversifying algorithm. The most notable feature of the proposed NMA is the use of randomized improvement-first local search instead of 2-opt. Our experiment on TSPLIB instances shows that while the populations evolved by our NMA tend to contain clusters at tight quality constraints, they frequently occupy distant basins of attraction rather than close-by regions, improving on the baseline diversification in terms of sum-sum diversity. Compared to the original NMA, ours, despite its simplicity, finds more distant solutions of higher quality within less running time, by a large margin. 
%With this comparison, we emphasize the impact of local search on NMA's performance in the context of diverse TSP solutions.
\end{abstract}
\keywords{Traveling Salesperson Problem, niching, evolutionary diversity optimization}
\section{Introduction}

% Evolutionary algorithms are bio-inspired metaheuristics that have found a wide range of applications in important areas such as combinatorial optimization, engineering and finance. They provide generic and modular frameworks to efficiently generate solutions to very difficult optimization problems, which are often of practical relevance \cite{Yar2016,Bianchi2008}.

Recently, research in optimization has seen rising interest in diverse solution problems, where multiple maximally distinct solutions of high quality are sought instead of a single solution \cite{Ingmar2020,Baste2020,hanaka2020finding}. This class of problem captures the need for multiple solutions and addresses many practical issues largely overlooked in traditional optimization. For instance, having diverse solutions provides robust alternatives that allow quick adaptation to changes in the problems rendering the current solution infeasible. Furthermore, it gives the users the flexibility to correct for gaps between the problem models and real-world settings, usually arising from errors in estimating model parameters, or when certain aspects of the problem cannot be expressed mathematically \cite{Schittekat2009}. On the other hand, diverse solution sets contain rich information about the problem instance (as opposed to similar solution sets), which the users can use to further augment their decision making capabilities. While there are methods to enumerate high quality solutions, having too many may not help the decision makers \cite{Glover2000}, and a small, diverse subset can be more manageable.

At its core, a diverse solution problem is an extension to an optimization problem, and it has been studied for many important problem classes in the last three decades. Studies on diverse solution problem to constraint satisfaction and optimization include interesting hardness results and introductions of various heuristics \cite{Emmanuel05,Petit15,Ruffini2019}. Promising attempts have been made to leverage existing solver frameworks to compute diverse solutions to SAT and Answer Set Problem \cite{Nadel2011,Eiter2009}. For Mixed Integer Programming, incorporation of diversity into quality-based heuristics has been investigated from various angles \cite{Glover2000,Danna,Trapp2015}. More recently, the first provably fixed-parameter tractable algorithms have been proposed for diverse solutions to a broad class of graph-based vertex problems \cite{Baste2020}, via modification of dynamic programming on the graph's tree decomposition. This inspired subsequent research that considers other combinatorial structures such as trees, paths \cite{hanaka2020finding}, matchings \cite{Fomin20211}, and independent sets \cite{Fomin2021}.

In evolutionary computation, there have been various works addressing the diverse solution problem from different angles. A popular search paradigm in robotics and content generation is quality diversity (QD) \cite{Pugh2016,Cully2018,Gravina2019,Alvarez2019}, which aims to illuminate the space of solution behaviors with a large population. Approaches in this paradigm are predicated on mapping solutions to behavior spaces, which typically demands domain expertise. On the other hand, evolutionary diversity optimization (EDO) focuses on the solution structures directly, and thus considers problems more closely matching with diverse solution problems as studied outside of evolutionary computation research. The idea of prioritizing diversity in genetic algorithms can be traced back to Ronald \cite{Ronald}, followed by more substantial works from Ulrich and Thiele \cite{Ulrich2010,Ulrich2011}. More recent studies investigate EDO for various cominatorial problems such as traveling salesperson problem (TSP) \cite{Do2020,Nikfarjam2021,Do2021,Nikfarjam20211}, knapsack \cite{Bossek2021}, minimum spanning tree \cite{Bossek20211}, and constrained subset selection \cite{Neumann2021}. It has also been explored in generating diverse TSP instances \cite{Alexander2017,Neumann2018,Neumann2019,Bossek2019,Gao2020}, useful in differentiating algorithms' performances.

Multimodal optimization, and more specifically evolutionary multimodal optimization also aims to find multiple solutions to optimization problems \cite{Shir2012,wong2015evolutionary,Preuss2015,Li2017}. This is a long-standing research area, having seen five decades of active interests. While most effort in this area deals with continuous optimization, relatively fewer works address multiple solutions generation in hard combinatorial problems. These include methods to find multiple TSP tours using genetic algorithm \cite{Ronald,Huang2018,Huang2019,Huang2020} and ant-colony optimization \cite{Angus2006,Han2018}. While algorithms in this area typically employ diversity promoting mechanisms such as speciation and crowding, they are not designed to address the diverse solution problem effectively since solution diversity only serves as the means to drive exploratory searches, and is not an objective in its own right. Nevertheless, by treating diversity as an end goal rather than just a feature of the search process, one may yet adopt these methods to achieve state-of-the-art performances on diverse solution problems.

In this work, we investigate a first application of a niching-based method from evolutionary multimodal optimization to finding diverse high quality TSP tours. The problem we consider is to find a fixed-size set of tours maximizing some diversity measure defined on edges, under a maximum tour cost constraint. As TSP is generally hard to approximate, satisfying the quality constraint can be difficult, and conventional ``anytime'' algorithm design no longer suffices. To address this issue, we modify and integrate a well-performing niching memetic algorithm (NMA) for multi-solution TSP \cite{Huang2019} into a 2-stage approach. The underlying local search heuristic in the algorithm helps handle the quality constraint with strong exploitation, while the niching strategy maintains some diversity in convergence paths. By combining exploratory search facilitated by niching, and a baseline diversification scheme $(\mu+1)$-EA in \cite{Do2020}, we retrofit an algorithm designed for multi-solution problems to address the diverse solution problem. Most modifications involve simplification of many procedures, and swapping the 2-opt local search for a randomized improvement-first version. The end result is a transparent algorithm that allows us to study the effectiveness of niching in solution diversification. Our experiment on 10 TSPLIB instances shows that the algorithm tends to discover distant regions of optima, achieving high sum-sum diversity as the result. On the other hand, the algorithm's outputs usually form clusters at tight quality constraints, leading to diversification reaching low sum-min diversity on some instances. Furthermore, our NMA, despite being simpler, significantly outperforms the original as it finds more distant solutions of higher quality, within less running time. The comparison highlights the importance of local search, where speed and diverse convergence paths are preferred over ability to reach the optimum.

This paper is organized as follows. Section \ref{sec:pre} includes the definition of the diverse TSP solutions problem, and a brief outline of $(\mu+1)$-EA for diversification. We introduce our 2-stage approach in Section \ref{sec:alg}, with a focus on the NMA. Section \ref{sec:exp} contains our experimental investigation and results. Finally, we conclude this study in Section \ref{sec:con}.

\section{Preliminaries}\label{sec:pre}

The symmetric TSP is formulated as follow. Given a complete undirected graph $G=(V,E)$ with $n=|V|$ vertices, $m=n(n-1)/2=|E|$ edges and the distance function $d:V\times V\to\mathbb{R}_{\geq 0}$, the goal is to compute a tour of minimal cost that visits each vertex exactly once and finally returns to the original node. A tour $I$ can be represented by the permutation $\pi:V\to V$, and we denote its edge set with $E(\pi)$ or $E(I)$. The TSP asks to find such $\pi$ that minimizes the tour cost

\begin{align*}
c(\pi) = d(\pi(n),\pi(1)) + \sum_{i=1}^{n-1} d(\pi(i),\pi(i+1)). 
\end{align*}

In this work, we consider diversity optimization for TSP. For each problem instance, we are to find a set $P$ of $\mu$ solutions that is diverse with respect to some diversity measure, while each solution meets a given quality threshold. Typically, such an instance requires every solution to be a $(1+\alpha)$-approximation to the underlying TSP instance, where $\alpha>0$. Here, we assume that the threshold value is provided to the algorithm instead of the gap ratio $\alpha$. Formally, given a TSP instance with the tour space $S$, the cost function $c$, and a diversity function $div$, we define a $(\mu,\alpha)$-instance of the diversity optimization problem as
\begin{equation}\label{eq:prob}
\max_{|P|=\mu}\left\lbrace div(P)|\forall I \in P,c(I)\geq(1+\alpha)\max_{J\in S}\{c(J)\}\right\rbrace.
\end{equation}

Here, we consider two diversity measures as no single measure captures all interesting aspects of diversity in a set. These measures are the normalized sum-sum and the sum-min scores, defined with $D_1$ and $D_2$, respectively,
\begin{equation}\label{eq:div_measure}
D_1(P)=\sum_{I,J\in P}\frac{dist(I,J)}{\mu(\mu-1)},\quad D_2(P)=\sum_{I\in P}\min_{J\in P\setminus\{I\}}\left\lbrace \frac{dist(I,J)}{\mu}\right\rbrace,
\end{equation}
where $dist(\cdot,\cdot)$ is the edge distance
\begin{equation}\label{eq:edge_dist}
dist(I,J)=1-\frac{|E(I)\cap E(J)|}{n}.
\end{equation}

%\frank{The next two paragraphs could become part of introduction as they set the scene and motivate your work}
This extension to the TSP is highly non-trivial due to additional considerations not present in traditional optimization. Since TSP is generally not efficiently approximable within a constant ratio \cite{Sahni1976} (although metric-TSP admits a $1.5$-FPTAS \cite{Chr76}, and Euclidean-TSP admits a $(1+\epsilon)$-PTAS \cite{Arora1998}), it is possible (and even probable) that a supposedly anytime algorithm, such as an evolutionary algorithm, returns an infeasible solution set when terminated during a run, that is a set containing at least a non-$(1+\alpha)$-approximation. One would consider this a failure case at first glance, since the algorithm fails to find distinct $\mu$ satisfactory solutions. However, if $\mu$ is greater than the number of all $(1+\alpha)$-approximations for an instance, then duplicates would be inevitable if one were to solve the problem as specified. Therefore, in this work, we accept duplicates, and regard an algorithm's run as failure only if no satisfactory solution is found. It is clear that the presence of duplicates diminishes diversity, and thus is discouraged via diversity maximization using appropriate diversity measures. On the other hand, this relaxation still does not guarantee the ``anytime'' assumption, so a somewhat unorthodox search algorithm design is called for.

Previously, a simple $(\mu+1)$-EA is considered for the problem \eqref{eq:prob}, which essentially performs random local search in the feasible population space \cite{Do2020}. The algorithm is outlined in Algorithm \ref{alg:ea}. It changes the population one solution at a time via a mutated offspring. The diversity measure used in the algorithm either makes the population disperse towards unused edges, or separate solutions from each other. This algorithm operates on the assumption that the initial population is feasible, and thus is an anytime algorithm. While this assumption no longer holds in our setting, the algorithm is still applicable if satisfactory solutions are found beforehand, using a different procedure. On the other hand, this problem is NP-hard even when starting with a feasible population, since it is equivalent to the dispersion problem over an unknown ground set, i.e. the set of all satisfactory solutions. The problem is known to be hard even with known ground sets and metric distance functions \cite{Wang1988,Erkut1990,Ravi1994,Chandra1996}.

\begin{algorithm}[t]
\begin{algorithmic}[1]
\STATE \textbf{Inputs:} TSP instance $c$, set size $\mu$, $threshold$, $initialPop$
\STATE $P\gets initialPop$
\WHILE{stopping criteria not met}
\itemindent=0pc
\STATE $I\gets UniformSample(P)$
\STATE $I'\gets Mutate(I)$
\IF{$c(I')\leq threshold$}
\itemindent=0pc
\STATE $P\gets P\cup\{I'\}$
\STATE $I''\gets\argmax_{J\in P}\{diversity(P\setminus\{J\})\}$
\STATE $P\gets P\setminus\{I''\}$
\ENDIF
\ENDWHILE
\STATE \textbf{return} $P$
\end{algorithmic}
\caption{$(\mu+1)$-EA for diversity optimization}
\label{alg:ea}
\end{algorithm}

Since Algorithm \ref{alg:ea} performs local search heuristics in the population space, it is prone to be stuck in local optima. The objective landscapes in TSP are typically multimodal when viewed over most reasonable neighborhood structures (e.g. ones defined by edge distance limits). This can be seen from the fact that k-opt, for all $k\in[2,n/2-5]$, fails to guarantee optimality even on metric-TSP instances \cite{Chandra94newresults}. Therefore, when a sufficiently tight objective threshold is imposed, the feasible solution space will be composed of disconnected regions. In such cases, we can see that Algorithm \ref{alg:ea} would fail to explore regions other than those that the initial population occupies, thus may never achieve the most diverse feasible population. As these regions correspond to a subset of local optima, this necessitates the use of an optima discovery mechanism,  and thus motivates our investigation of niching strategies.

\section{Diversity maximization with Niching memetic algorithm}\label{sec:alg}
%\frank{Divide into subsections explaining the different modules}
In this work, we propose a 2-stage search approach: the first stage is dedicated to finding a sufficient number of satisfactory solutions, and the second focuses on diversification. The whole procedure is outlined in Algorithm \ref{alg:all}, where the first stage is at line \ref{line:nma} and the second at line \ref{line:diversify}. The procedure for the second stage is described in Algorithm \ref{alg:ea}. We note that this is similar to the existing 2-stage EDO strategies for subset selection \cite{Neumann2021} and knapsack \cite{Bossek2021}, where the first stage deals with the quality constraint with an approximation algorithm. This section focuses on the algorithm used in the first stage.

\begin{algorithm}[t]
\begin{algorithmic}[1]
\STATE \textbf{Inputs:} TSP instance $c$, set size $\mu$, $threshold$
\STATE $P\gets NMA(c,\mu,threshold)$\label{line:nma}
\IF{$|P|>\mu$}
\itemindent=0pc
\STATE $P\gets GMM(P,\mu)$
\ENDIF
\STATE \textbf{return} $(\mu+1)$-$EA(c,\mu,threshold,P)$\label{line:diversify}
\end{algorithmic}
\caption{2-stage EDO approach}
\label{alg:all}
\end{algorithm}

It is important that the result of the first stage be a good seed for the second stage, meaning it should occupy as many feasible regions as possible. To this end, we rely on an existing approach as a basis for our design. An effective niching memetic algorithm for multi-solution TSP was proposed and shown to perform well on small benchmark instances in terms of finding global optima \cite{Huang2019}. Aside from its performance, it is also relatively simple, allowing us to isolate and study the effect of niching on diversification performance. That said, modifications to the algorithm are needed, as the problem it was designed to solve is different from the one we consider. The result is outlined in Algorithm \ref{alg:nma}, which is used to execute the first stage. The output of this procedure is the initial population for Algorithm \ref{alg:ea} in the second stage. However, if it contains more than $\mu$ solutions, an approximation heuristic called GMM (viz. Greedy heuristic for Maximizing Minimum distance) \cite{Ravi1994} is applied to select a diverse $\mu$-subset, as efficiency is prioritized over optimality at this point. This heuristic uses the edge distance defined in \eqref{eq:edge_dist}, starts with a set of the two furthest elements, iteratively adds the furthest element from the set, and terminates after adding enough elements.

\begin{algorithm}[h]
\begin{algorithmic}[1]
\STATE \textbf{Inputs:} TSP instance $c$, set size $\mu$, $threshold$
\STATE $P\gets Initialize(c,\mu)$
\STATE Evaluate $P$
\WHILE{$|\{I\in P,c(I)\leq threshold\}|<\mu$}
\itemindent=0pc
\STATE $Grouping\gets NeighborhoodStrategy(P,c,M_{min},M_{max})$
\STATE $Grouping\gets DiversityEnhancement(P,Grouping)$
\STATE $Off\gets Crossover(P,Grouping)$
\STATE $Off\gets Mutate(Off)$
\STATE $SelectiveEvaluate(Off,c)$
\STATE $Off\gets LocalSearch(Off,c,Grouping,threshold)$
\STATE $P\gets Replacement(P,Off,Grouping)$
\ENDWHILE
\STATE \textbf{return} $\{I\in P,c(I)\leq threshold\}$
\end{algorithmic}
\caption{Niching Memetic Algorithm for EDO}
\label{alg:nma}
\end{algorithm}

First, we outline the aspects of the algorithm that are unchanged from the original design \cite{Huang2019}. For population initialization, each solution is created by sampling the first $n/2$ vertices, and filling the rest with greedy vertex selection. Then, the population is divided into groups using the original adaptive neighborhood strategy, where the group sizes are controlled by the leaders' cost; the leader is the best solution in each group. For crossover, PMX \cite{Goldberg1985} is used due to its low computational cost, and ability to introduce many edges not in the parents. For mutation, the 4-edge-exchange via swapping two vertices is used as it may lead to branching out to different convergence paths from applying 2-edge-exchange local search. Additionally, all parameters used in these components are kept the same.

\subsection{Simplification}

A major difference between our design in the original is the lack of the critical edge set, which is used to store edges frequently appearing in groups' leaders. Such edges are supposed to approximate the common edges shared by good solutions, and thus should be kept in the population's edge pool. This is done by forcing the variation operators to avoid removing such edges, which also supposedly reduces unnecessary exploitation. However, in our problem, these edges would quickly become over-represented in the population, diminishing diversity. Leaders of different groups would likely converge to the same optima as the result of maintaining common edges. Furthermore, the efficiency of the main exploitation mechanism, i.e. the local search, can be improved significantly in a different manner, which we will describe shortly.

\subsection{Diversity enhancement with migration}

Another change is in the diversity enhancement heuristic used to keep each group from stagnation due to genetic drift. Originally, when a group fully converges such that it only contains duplicates, non-leaders are mutated so as to explore the nearby area. We opt for a more aggressive migration-based approach where each duplicate of the leader is swapped with a random non-leader from a different group. The procedure is outlined in Algorithm \ref{alg:divEnhance}. For the highly converging group, this introduces new genetic materials which improve the effectiveness of crossover and, in turn, the range of exploration. For the other group, this provides potentially good genes which facilitate in-group competitiveness and exploitation. Since each duplicate is migrated to a random group, these duplicates will likely be moved to different groups, so it is unlikely that their genes will dominate the affected group's gene pool, given the exploitation strength of the local search. Note that this heuristic modifies no solution, so re-evaluation is not necessary at this point. It also does not guarantee that a duplicate is moved out of the group, as it may be subsequently swapped back, though this is very unlikely based on our observations.

\begin{algorithm}[t]
\begin{algorithmic}[1]
\STATE \textbf{Inputs:} $P$, $Grouping$
\STATE $G\gets Grouping$
\FOR{$g$ in $G$}
\itemindent=0pc
\STATE $L\gets P(g.leader)$
\FOR{non-leader $i\in g$}
\itemindent=0pc
\IF{$dist(P(i),L)=0$}
\itemindent=0pc
\STATE $g'\gets UniformSample(G\setminus\{g\})$
\STATE $i'\gets UniformSampleNonLeader(g')$
\STATE $g\gets (g\setminus\{i\})\cup\{i'\}$
\STATE $g'\gets (g'\setminus\{i'\})\cup\{i\}$
\ENDIF
\ENDFOR
\ENDFOR
\STATE \textbf{return} $G$
\end{algorithmic}
\caption{DiversityEnhancement}
\label{alg:divEnhance}
\end{algorithm}

\subsection{Local search}

The largest change is in the local search heuristic. Greedy local search, i.e. 2-opt, is used in the original design, where the best out of $n(n-3)/2$ 2-edge-exchanges is chosen in each step. Instead, we use improvement-first strategy where an inversion is performed whenever an improvement is found during the scan through the 2-opt neighborhood. This allows multiple improvements in a single pass, leading to a much faster boost to the solution's quality. Furthermore, while greedy local search usually yields solutions closer to optimal quality, our algorithm only needs to find those satisfying a given quality threshold, so improvement-first is effective. On that note, our implementation terminates the local search procedure when the threshold is reached, even before reaching a local optimum. This reduces unnecessary exploitation, and mitigates diversity loss. Additionally, to eliminate positional biases in the improvement-first strategy, the neighborhood is scanned in a random order each time, by making use of a random permutation over $[n(n-3)/2]$. Such randomness in an otherwise deterministic procedure could also induce some diversity in convergence paths. Also, we observed that this heuristic often causes quick convergence, so we set a limit of $4(n-3)$ evaluations on each local search call for a solution. We found that giving each local search call more budget often slows down the process of finding multiple satisfactory solutions, and frequent propagation of good genes from local search via crossover and selection is more effective. Note that we consider each 2-opt neighbor lookup to cost $4/n$ evaluation, same as in \cite{Huang2019}.

Similar to the original design, only the best half of each group (rounded up) undergo local search. The key difference is that our version excludes some solutions from the pool to be considered for local search. These are solutions satisfying the cost threshold, and those known to be local optima. The choice of the former is obvious considering the stopping criteria of the local search heuristic. In this scheme, if more than half the group are excluded, fewer solutions will undergo local search. Here, a solution is deemed a local optimum if it was last altered by local search, which terminated upon finding no improvement within a full pass through the neighborhood.

\subsection{Selection}

As for the survival selection, we opt for a simpler procedure. As per the speciation strategy, competition is contained within groups. In the original design, a crowding strategy is used where each offspring solution competes with the nearest parent in the group, and with other offspring in the group that compete with the same parent. The best one within this tournament replaces the parent. This approach tends to waste computational effort from local search, as it often drives the offpsring far away from parents, leading to good converging offspring eliminating each other. Instead, we use group-wide elitism, and rely on the aforementioned diversity enhancement heuristic and truncating the local search to maintain diversity. Since the second stage of our approach only starts when the niching strategy ends, faster convergence afforded by elitism leads to more time to diversify the population later on.

\subsection{Post-processing}

Since our approach is designed for a different problem, the solution preservation procedure needs to be changed. In the original design, only the best solutions in each group are considered for keeping, and added to the output set in decreasing quality order (i.e. increasing cost). During this procedure, each of them is discarded if its quality is not within a small gap from the best found, or if it is too close to an previously added solution. This approach is not suitable for extracting a predetermined number of solutions, and requires a large population, which decreases the efficiency of exploitation. Instead, our approach simply returns all satisfactory solutions, which are subjected to diversification in the second stage. Such simplicity gives the transparency needed to gain insight into the performance of niching in our approach.

\subsection{Complexity}

In terms of complexity of Algorithm \ref{alg:nma}, most computational effort is from the local search, though this diminishes as the population approaches local optima or the quality threshold. Another costly operation is the distance calculation in the neighborhood strategy, with time complexity scaling quadratically w.r.t. the population size. We remark that this is comparable to the overhead in Algorithm \ref{alg:ea} from using and updating the distance or count table. Moreover, in order to find $\mu$ solutions for the second stage, the population size in the first stage must be at least $\mu$. Therefore, the time complexity per step is much higher in Algorithm \ref{alg:nma} than in Algorithm \ref{alg:ea}. Still, the simplified subroutines and the selective evaluations of offspring contribute to lower overall complexity than that of the original design.

\section{Experiments}\label{sec:exp}

In the experiments, we use 10 TSPLIB 2D-Euclidean instances of varying sizes for benchmark. Here, we aim to investigate the outputs from our NMA (Algorithm \ref{alg:nma}), and how they augment the performance of $(\mu+1)$-EA, (Algorithm \ref{alg:ea}) when used as the initial populations. Additionally, we compare these results to those from the original niching memetic algorithm, both with and without the second stage.

\subsection{Experimental setup}

For each instance of $n$ vertices, we consider the required solution set size $\mu=\lfloor n/4\rfloor$, and the cost thresholds corresponding to gaps $\alpha\in\{0.05,0.1,0.2\}$. For the first stage with NMA, we set the parameters similar to those in \cite{Huang2019}: crossover rate $0.9$, mutation rate $0.01$, group size range $[4,12]$. We set the population size to be $3\mu$. For the second stage with $(\mu+1)$-EA, we use both variants \cite{Do2020}: equalizing edge frequency (ED), and maximizing pairwise distances (PD). These variants are run with the same initial population from the NMA in each run to facilitate better comparison, and we denote these with NMA-ED and NMA-PD, respectively. The evaluation budget shared by both stages is $40\lfloor \mu n\sqrt{n}\rfloor$, and $30$ runs are done independently on each instance. We note that the ED and PD variants maximize $D_1$ and $D_2$ in Eq \eqref{eq:div_measure}, respectively.

To establish baseline diversity scores, we run the two variants of $(\mu+1)$-EA with duplicates of the optimal tour as the initial population. It is run with the same evaluation budget as above, so the entire running time is spent on diversification. This allows us to observe the effectiveness of NMA in the overall procedure\footnote{The implementation is available at \url{https://github.com/DV-Anh/EDO-Niching}}.

To compare against the original NMA, we run the source code\footnote{\url{https://github.com/GnauhGnit/MSTSP}} provided by the authors, with the same parameter setting as above. In the preservation procedure, we set $\alpha$ as the filtering gap. We also run the algorithm separately with population size $150$ to mimic the setting in \cite{Huang2019}, so as to better expose the impact of our modifications. Note that on 9 out of 10 instances, the evaluation budgets are significantly smaller than those in \cite{Huang2019}, so we expect worse results than reported there. Also, since their and our versions of NMA are designed for different problems, we are only interested in comparing the convergence performances and numbers of found distant solutions, not diversity scores.

\subsection{Diversity and running time results}

We collect the $D_1$ and $D_1$ scores (Eq. \eqref{eq:div_measure}) of the final populations and report the means in Table \ref{tab:results_div}. Kruskal-Wallis tests with Bonferroni correction \cite{Corder09} are performed on these scores, with the results shown in Table \ref{tab:results_stat}. We can see that the niching method makes the greatest impact when $\alpha$ is small, and the differences diminish at larger thresholds. This aligns with our assumption that the feasible regions are more fragmented at a tighter constraint, leading to limited diversity from $(\mu+1)$-EA. It can be seen that at $\alpha=0.05$, NMA-ED and NMA-PD mostly achieve higher $D_1$ scores than their non-NMA counterparts, within less diversification time. This shows that NMA outputs populations that occupy multiple distant regions, resulting wider coverage of the solution space. On the other hand, the similar $D_1$ scores at larger thresholds suggest that more connected feasible regions pose less of a challenge to a simple local search heuristic, especially when seeded with the optimal solution.

In terms of $D_2$, the differences are more nuanced. At $0.05$ and $0.1$ values of $\alpha$, the populations from NMA-PD have higher $D_2$ scores than those from PD only up to kroA100, and the differences are reversed in larger instances. Meanwhile, NMA-PD maintains higher $D_1$ scores than PD at $\alpha=0.2$ across all instances, mostly with statistical significance. On the other hand, NMA-ED achieves lower $D_2$ scores than ED in most cases, albeit with only few statistical significance cases, observable only at $\alpha=0.05$. This indicates that the outputs from the niching method consist of converged clusters that are far from each other, according to the high $D_1$ scores. This property seems to be a direct consequence of grouping in the niching method. The low $D_2$ scores, compared to those of PD, might indicate the small size of the feasible region each cluster occupies, compared to the one the optimal solution occupies.

\begin{table*}[t]
\centering
\caption{Mean of $D_1$ and $D_2$ values from 4 algorithms.}
\label{tab:results_div}
\begin{tabular}{llcccccccc}\toprule
\multicolumn{1}{c}{\multirow{2}{*}{Instance}} &\multicolumn{1}{c}{\multirow{2}{*}{$\alpha$}} &\multicolumn{2}{c}{ED (1)}&\multicolumn{2}{c}{PD (2)}&\multicolumn{2}{c}{NMA-ED (3)}&\multicolumn{2}{c}{NMA-PD (4)}\\\cmidrule(l{2pt}r{2pt}){3-10}
&&\multicolumn{1}{c}{$D_1$} &\multicolumn{1}{c}{$D_2$} &\multicolumn{1}{c}{$D_1$} &\multicolumn{1}{c}{$D_2$} &\multicolumn{1}{c}{$D_1$} &\multicolumn{1}{c}{$D_2$} &\multicolumn{1}{c}{$D_1$} &\multicolumn{1}{c}{$D_2$} \\\midrule
\multirow{3}{*}{eil51}&0.05&34.949\%&12.102\%&32.479\%&26.062\%&41.778\%&12.881\%&41.559\%&32.729\%\\
&0.1&51.407\%&12.663\%&48.457\%&43.279\%&52.894\%&16.035\%&53.264\%&48.568\%\\
&0.2&66.313\%&33.508\%&64.355\%&60.997\%&66.725\%&34.875\%&65.868\%&62.941\%\\
\midrule \multirow{3}{*}{berlin52}&0.05&35.088\%&10.232\%&33.692\%&29.453\%&36.942\%&8.496\%&36.168\%&29.004\%\\
&0.1&49.326\%&18.195\%&47.333\%&43.590\%&50.027\%&14.329\%&49.656\%&44.103\%\\
&0.2&65.308\%&34.606\%&63.037\%&60.000\%&65.188\%&32.865\%&64.994\%&61.977\%\\
\midrule \multirow{3}{*}{st70}&0.05&37.003\%&10.291\%&36.280\%&31.493\%&41.195\%&6.395\%&40.585\%&33.109\%\\
&0.1&48.476\%&11.714\%&46.984\%&42.412\%&50.459\%&10.700\%&50.320\%&45.782\%\\
&0.2&62.133\%&23.339\%&61.181\%&58.076\%&62.503\%&23.686\%&61.768\%&58.902\%\\
\midrule \multirow{3}{*}{eil76}&0.05&31.121\%&5.674\%&29.992\%&24.451\%&39.175\%&3.966\%&39.142\%&27.604\%\\
&0.1&48.078\%&8.167\%&45.950\%&41.055\%&50.170\%&8.287\%&51.099\%&46.690\%\\
&0.2&64.807\%&20.025\%&62.420\%&59.007\%&65.152\%&20.321\%&65.051\%&62.211\%\\
\midrule \multirow{3}{*}{kroA100}&0.05&31.066\%&5.459\%&29.823\%&24.051\%&34.119\%&3.511\%&32.950\%&24.821\%\\
&0.1&45.206\%&8.123\%&42.990\%&38.072\%&46.113\%&6.089\%&44.030\%&38.773\%\\
&0.2&59.324\%&18.744\%&57.390\%&53.727\%&59.067\%&17.004\%&58.143\%&54.835\%\\
\midrule \multirow{3}{*}{eil101}&0.05&35.542\%&5.488\%&33.638\%&28.816\%&40.722\%&3.127\%&40.222\%&24.289\%\\
&0.1&50.959\%&7.962\%&48.002\%&43.842\%&52.211\%&7.042\%&53.193\%&48.185\%\\
&0.2&66.112\%&16.669\%&64.024\%&60.804\%&66.524\%&18.912\%&66.524\%&63.979\%\\
\midrule \multirow{3}{*}{lin105}&0.05&30.839\%&4.284\%&29.867\%&24.131\%&32.235\%&3.520\%&30.007\%&22.348\%\\
&0.1&43.363\%&6.530\%&43.066\%&37.600\%&43.212\%&4.426\%&41.192\%&35.217\%\\
&0.2&57.497\%&14.241\%&57.413\%&53.750\%&57.420\%&13.681\%&56.728\%&53.005\%\\
\midrule \multirow{3}{*}{ch150}&0.05&27.206\%&3.264\%&26.120\%&21.146\%&31.998\%&1.828\%&30.800\%&18.606\%\\
&0.1&41.525\%&4.309\%&39.080\%&34.823\%&42.007\%&2.882\%&40.918\%&34.513\%\\
&0.2&56.648\%&7.809\%&54.329\%&50.957\%&56.577\%&6.377\%&55.683\%&52.511\%\\
\midrule \multirow{3}{*}{tsp225}&0.05&27.597\%&2.117\%&25.334\%&20.745\%&31.891\%&1.146\%&29.682\%&11.800\%\\
&0.1&43.393\%&2.741\%&39.375\%&34.903\%&45.358\%&1.928\%&44.021\%&33.393\%\\
&0.2&59.086\%&5.141\%&56.105\%&52.634\%&59.034\%&5.357\%&58.715\%&55.785\%\\
\midrule \multirow{3}{*}{pcb442}&0.05&30.301\%&1.451\%&27.691\%&23.350\%&32.757\%&0.680\%&30.138\%&14.340\%\\
&0.1&43.382\%&1.538\%&39.854\%&36.316\%&42.762\%&1.094\%&41.734\%&33.402\%\\
&0.2&56.414\%&1.961\%&53.447\%&50.810\%&55.314\%&1.598\%&54.238\%&51.808\%\\
\bottomrule
\end{tabular}
\end{table*}

\begin{table}[t]
\centering
\caption{Statistical test results on $D_1$ and $D_2$}
\label{tab:results_stat}
\small
\begin{tabular}{llll}
\toprule
\multicolumn{1}{c}{Instance} & \multicolumn{1}{c}{$\alpha$} & \multicolumn{1}{c}{$D_1$} & \multicolumn{1}{c}{$D_2$} \\ \midrule
\multirow{3}{*}{eil51}&0.05&1\textless3;1\textless4;2\textless3;2\textless4&1\textless2;1\textless4;2\textgreater3;2\textless4;3\textless4\\
&0.1&1\textgreater2;1\textless3;1\textless4;2\textless3;2\textless4&1\textless2;1\textless4;2\textgreater3;2\textless4;3\textless4\\
&0.2&1\textgreater2;2\textless3;2\textless4;3\textgreater4&1\textless2;1\textless4;2\textgreater3;2\textless4;3\textless4\\
\midrule \multirow{3}{*}{berlin52}&0.05&1\textgreater2;1\textless3;1\textless4;2\textless3;2\textless4&1\textless2;1\textless4;2\textgreater3;3\textless4\\
&0.1&1\textgreater2;2\textless3;2\textless4&1\textless2;1\textless4;2\textgreater3;3\textless4\\
&0.2&1\textgreater2;2\textless3;2\textless4&1\textless2;1\textless4;2\textgreater3;2\textless4;3\textless4\\
\midrule \multirow{3}{*}{st70}&0.05&1\textless3;1\textless4;2\textless3;2\textless4&1\textless2;1\textgreater3;1\textless4;2\textgreater3;3\textless4\\
&0.1&1\textgreater2;1\textless3;1\textless4;2\textless3;2\textless4&1\textless2;1\textless4;2\textgreater3;2\textless4;3\textless4\\
&0.2&1\textgreater2;2\textless3;2\textless4;3\textgreater4&1\textless2;1\textless4;2\textgreater3;3\textless4\\
\midrule \multirow{3}{*}{eil76}&0.05&1\textless3;1\textless4;2\textless3;2\textless4&1\textless2;1\textless4;2\textgreater3;3\textless4\\
&0.1&1\textgreater2;1\textless3;1\textless4;2\textless3;2\textless4;3\textless4&1\textless2;1\textless4;2\textgreater3;2\textless4;3\textless4\\
&0.2&1\textgreater2;2\textless3;2\textless4&1\textless2;1\textless4;2\textgreater3;2\textless4;3\textless4\\
\midrule \multirow{3}{*}{kroA100}&0.05&1\textless3;1\textless4;2\textless3;2\textless4&1\textless2;1\textless4;2\textgreater3;3\textless4\\
&0.1&1\textgreater2;1\textgreater4;2\textless3;2\textless4;3\textgreater4&1\textless2;1\textless4;2\textgreater3;3\textless4\\
&0.2&1\textgreater2;1\textgreater4;2\textless3;2\textless4;3\textgreater4&1\textless2;1\textless4;2\textgreater3;2\textless4;3\textless4\\
\midrule \multirow{3}{*}{eil101}&0.05&1\textgreater2;1\textless3;1\textless4;2\textless3;2\textless4&1\textless2;1\textless4;2\textgreater3;2\textgreater4;3\textless4\\
&0.1&1\textgreater2;1\textless3;1\textless4;2\textless3;2\textless4;3\textless4&1\textless2;1\textless4;2\textgreater3;2\textless4;3\textless4\\
&0.2&1\textgreater2;1\textless3;1\textless4;2\textless3;2\textless4&1\textless2;1\textless4;2\textgreater3;2\textless4;3\textless4\\
\midrule \multirow{3}{*}{lin105}&0.05&1\textgreater2;1\textless3;1\textgreater4;2\textless3;3\textgreater4&1\textless2;1\textless4;2\textgreater3;3\textless4\\
&0.1&1\textgreater4;2\textgreater4;3\textgreater4&1\textless2;1\textless4;2\textgreater3;2\textgreater4;3\textless4\\
&0.2&1\textgreater4;2\textgreater4;3\textgreater4&1\textless2;1\textless4;2\textgreater3;3\textless4\\
\midrule \multirow{3}{*}{ch150}&0.05&1\textless3;1\textless4;2\textless3;2\textless4&1\textless2;1\textless4;2\textgreater3;3\textless4\\
&0.1&1\textgreater2;2\textless3;2\textless4;3\textgreater4&1\textless2;1\textless4;2\textgreater3;3\textless4\\
&0.2&1\textgreater2;1\textgreater4;2\textless3;2\textless4;3\textgreater4&1\textless2;1\textless4;2\textgreater3;2\textless4;3\textless4\\
\midrule \multirow{3}{*}{tsp225}&0.05&1\textgreater2;1\textless3;1\textless4;2\textless3;2\textless4;3\textgreater4&1\textless2;1\textgreater3;1\textless4;2\textgreater3;2\textgreater4;3\textless4\\
&0.1&1\textgreater2;1\textless3;2\textless3;2\textless4;3\textgreater4&1\textless2;1\textless4;2\textgreater3;3\textless4\\
&0.2&1\textgreater2;1\textgreater4;2\textless3;2\textless4&1\textless2;1\textless4;2\textgreater3;2\textless4;3\textless4\\
\midrule \multirow{3}{*}{pcb442}&0.05&1\textgreater2;1\textless3;2\textless3;2\textless4;3\textgreater4&1\textless2;1\textgreater3;1\textless4;2\textgreater3;2\textgreater4;3\textless4\\
&0.1&1\textgreater2;1\textgreater3;1\textgreater4;2\textless3;2\textless4;3\textgreater4&1\textless2;1\textless4;2\textgreater3;2\textgreater4;3\textless4\\
&0.2&1\textgreater2;1\textgreater3;1\textgreater4;2\textless3;2\textless4;3\textgreater4&1\textless2;1\textless4;2\textgreater3;2\textless4;3\textless4\\
\bottomrule
\end{tabular}
\end{table}

For further analysis, we inspect the average evaluations until the 2-stage approach stops improving diversity. These are reported in Table \ref{tab:results_runtime}, along with the average termination times of the first stage, NMA. Noticeably, ED, PD, NMA-ED, NMA-PD reach plateaus within budgets at small $\alpha$ and on smaller instances, whereas runs on larger instances show no sign of stopping the diversity increases. Whenever there are signs of plateaus, i.e. stopping improving within 80\% of budget, the NMA-ED and NMA-PD always achieve higher diversity scores than ED and PD, confirming that the niching strategy explores regions unreachable by local search. Additionally, the NMA always terminates well within the budgets, but uses significantly more of the budgets on larger instances than smaller ones (at $\alpha=0.05$), especially pcb442 at 62\% on average. This increases can be seen when $\alpha=0.1$, at a much lower rate. It indicates a scaling issue with NMA, and suggests that a more powerful local search heuristic would be very beneficial. The comparison to the original NMA in the next section further emphasizes the impact of local search on NMA's performance.

\begin{table}[t]
\centering
\caption{Mean time-till-plateau (until the last diversity improvement) of all variants, and mean times before termination of NMA, shown in percentages of the corresponding evaluation budgets.}
\label{tab:results_runtime}
\small
\setlength{\tabcolsep}{4.6pt}
\begin{tabular}{lcccccc}
\toprule
Instance & $\alpha$ & ED & PD & NMA & NMA-ED & NMA-PD \\ \midrule
\multirow{3}{*}{eil51}&0.05&63.199\%&74.927\%&9.239\%&44.718\%&45.342\%\\
&0.1&79.189\%&86.289\%&3.581\%&68.559\%&57.290\%\\
&0.2&90.114\%&93.895\%&1.443\%&89.768\%&77.561\%\\
\midrule \multirow{3}{*}{berlin52}&0.05&59.115\%&76.151\%&15.794\%&53.178\%&70.692\%\\
&0.1&90.085\%&89.464\%&5.947\%&70.789\%&70.677\%\\
&0.2&94.132\%&95.519\%&1.788\%&88.851\%&89.857\%\\
\midrule \multirow{3}{*}{st70}&0.05&73.839\%&90.381\%&10.885\%&66.362\%&87.161\%\\
&0.1&87.603\%&95.660\%&4.817\%&80.318\%&91.269\%\\
&0.2&96.707\%&98.554\%&1.576\%&94.672\%&97.123\%\\
\midrule \multirow{3}{*}{eil76}&0.05&78.947\%&92.381\%&17.681\%&66.658\%&77.438\%\\
&0.1&89.389\%&97.947\%&4.756\%&83.934\%&88.592\%\\
&0.2&98.296\%&99.059\%&1.330\%&97.852\%&96.415\%\\
\midrule \multirow{3}{*}{kroA100}&0.05&85.269\%&97.382\%&11.398\%&83.856\%&92.682\%\\
&0.1&93.999\%&99.435\%&5.404\%&88.697\%&98.302\%\\
&0.2&98.751\%&99.490\%&2.502\%&97.962\%&99.355\%\\
\midrule \multirow{3}{*}{eil101}&0.05&90.877\%&98.755\%&22.501\%&72.627\%&96.179\%\\
&0.1&99.117\%&99.458\%&4.508\%&94.222\%&97.412\%\\
&0.2&99.749\%&99.721\%&1.167\%&99.528\%&98.959\%\\
\midrule \multirow{3}{*}{lin105}&0.05&87.420\%&95.628\%&13.118\%&81.191\%&93.815\%\\
&0.1&94.504\%&99.353\%&5.516\%&92.896\%&97.660\%\\
&0.2&98.291\%&99.615\%&2.504\%&98.982\%&99.488\%\\
\midrule \multirow{3}{*}{ch150}&0.05&93.900\%&99.383\%&24.152\%&94.241\%&98.744\%\\
&0.1&98.274\%&99.776\%&7.486\%&96.482\%&99.303\%\\
&0.2&99.805\%&99.853\%&2.180\%&99.753\%&99.712\%\\
\midrule \multirow{3}{*}{tsp225}&0.05&98.927\%&99.819\%&48.802\%&99.702\%&99.694\%\\
&0.1&99.862\%&99.907\%&8.413\%&99.857\%&99.673\%\\
&0.2&99.964\%&99.951\%&1.784\%&99.959\%&99.903\%\\
\midrule \multirow{3}{*}{pcb442}&0.05&99.972\%&99.982\%&62.129\%&99.984\%&99.974\%\\
&0.1&99.989\%&99.986\%&11.441\%&99.983\%&99.969\%\\
&0.2&99.995\%&99.986\%&1.866\%&99.996\%&99.978\%\\
\bottomrule
\end{tabular}
\end{table}

\subsection{Comparing with the original NMA}

We collect the numbers of found solutions and the average gaps from the optimum, resulted from running the original NMA. For fair comparison, we further filter the final populations from our NMA (first stage only), NMA-ED and NMA-PD. To match the preservation heuristic in the original design, which ensures pairwise distance of at least $0.2$ in the output, we perform bottom-up single-linkage clustering with the same cutoff distance. The resulted number of clusters is then a lower bound of the number of solutions in the population, whose pairwise distances are at least $0.2$. We report these numbers, along with corresponding statistics from the original NMA, in Table \ref{tab:results_nma_comp}.

We see that in all instances, the original NMA delivers significantly fewer distant solutions compared to our approach, even without the diversification stage. This is likely due to the overly aggressive filtering in the preservation procedure. At smaller population sizes, the original NMA produces better solutions, but fewer of them. In most cases, it fails to meet the quality constraint, with average quality gaps exceeding $\alpha$. The results show that our NMA produces much better trade-offs between solution quality and distant solutions amount. Additionally, the quality gaps produced by the original NMA increase at larger instances. This, combines with the observation that our NMA uses more of the evaluation budgets in these instances, indicates a scaling issue with both versions. Nevertheless, ours demonstrates much improved efficiency by returning better solutions in greater amounts well within the budgets, whereas the original algorithm exhausts the budgets.

We observe that the local search heuristic plays a key role in NMA's performance. The first-improvement strategy seems to vastly boost the convergence speed of the procedure, especially considering that an additional tight evaluation budget is imposed on each local search call in our NMA, reducing the relative time spent on local search in each generation. While our NMA also terminates local search when the quality threshold is reached, the original NMA hardly even reaches this threshold within the budget most of the time, so ours must find better solutions for this termination criterion to even be effective. Meanwhile, the elitist survival selection could have increased the overall effectiveness of local search as most of improvements from local search tends to be carried over to the next generation. On the other hand, scanning through the neighborhood in random orders allows similar solutions to converge to different local optima. By stopping local search early, this divergence property is maintained, increasing the effectiveness of optima discovery. As it stands, fast improvement and diverse convergence paths seem to be more desirable local search properties in the diverse solution problem than reaching high quality optima.

\begin{table*}[t]
\centering
\caption{Comparison of numbers of output solutions and mean cost gaps w.r.t. the optimum. For fair comparison, hierarchical clustering was performed on the final populations from the new methods with cutoff distance of 0.2, and the numbers of clusters were reported.}
\label{tab:results_nma_comp}
\begin{tabular}{lccccccccccc}
\toprule
\multicolumn{1}{c}{\multirow{2}{*}{Instance}} &\multicolumn{1}{c}{\multirow{2}{*}{$\alpha$}} & \multicolumn{2}{c}{NMA (150)} & \multicolumn{2}{c}{NMA ($3\mu$)} & \multicolumn{2}{c}{NMA (modified)} & \multicolumn{2}{c}{NMA-ED} & \multicolumn{2}{c}{NMA-PD} \\
\cmidrule(l{2pt}r{2pt}){3-12}
&&\multicolumn{1}{c}{Gap} &\multicolumn{1}{c}{No. sol.} &\multicolumn{1}{c}{Gap} &\multicolumn{1}{c}{No. sol.} &\multicolumn{1}{c}{Gap} &\multicolumn{1}{c}{No. clus.} &\multicolumn{1}{c}{Gap} &\multicolumn{1}{c}{No. clus.} &\multicolumn{1}{c}{Gap} &\multicolumn{1}{c}{No. clus.} \\\midrule
\multirow{3}{*}{eil51}&0.05&1.3193&1.9&1.07&1.8333&1.0434&10.5667&1.0484&7.2&1.0485&11.6667\\
&0.1&1.3531&3.3667&1.0857&2.9333&1.0911&17.0333&1.0982&7.0667&1.0982&12\\
&0.2&1.4275&7.5&1.1181&3.8667&1.1904&17.9&1.1982&10.9333&1.1976&12\\
\midrule \multirow{3}{*}{berlin52}&0.05&1.2896&2.4333&1.0769&2&1.0433&5.1&1.0492&5.9667&1.0492&12.4667\\
&0.1&1.3077&3.9333&1.0991&3&1.0927&14.7&1.0993&7.4333&1.0991&13\\
&0.2&1.3835&7.9333&1.1149&4&1.1894&17.3667&1.199&11.8667&1.1987&13\\
\midrule \multirow{3}{*}{st70}&0.05&1.4068&1.9667&1.1318&1.7&1.0438&10.3333&1.0494&7.2&1.0493&16.8\\
&0.1&1.4404&3.1333&1.1675&2.5333&1.0924&21.5&1.0992&7.6667&1.099&17\\
&0.2&1.5482&6.6&1.2004&4.0667&1.1917&21.9667&1.1988&12.8333&1.1987&17\\
\midrule \multirow{3}{*}{eil76}&0.05&1.362&2.1333&1.1355&1.7&1.0455&11.2&1.0492&7.4&1.0491&17.1\\
&0.1&1.4173&4.1333&1.1831&3.4333&1.0947&24.8&1.0991&7.7333&1.099&19\\
&0.2&1.4596&8.1333&1.2151&4.9&1.1939&28.4333&1.1989&13.1667&1.1987&19\\
\midrule \multirow{3}{*}{kroA100}&0.05&1.5371&1.8333&1.3081&1.6333&1.0449&5.0667&1.0497&9.0667&1.0496&22.8333\\
&0.1&1.6055&3&1.3399&2.7333&1.0937&28.2&1.0995&8.3667&1.0994&25\\
&0.2&1.6869&6.9333&1.3733&4.5&1.1926&40.9333&1.1992&15.1333&1.1991&25\\
\midrule \multirow{3}{*}{eil101}&0.05&1.4038&2.0667&1.2053&1.6&1.0467&11.8333&1.0496&9.0667&1.0495&20\\
&0.1&1.4546&4.3&1.2501&3.4&1.0962&28.1&1.0994&8.8&1.0994&25\\
&0.2&1.5432&8.2&1.2928&5.9333&1.1953&37.4333&1.1992&16.1667&1.1991&25\\
\midrule \multirow{3}{*}{lin105}&0.05&1.5417&1.8667&1.2692&1.2333&1.0448&1.7&1.0497&10&1.0496&22.1667\\
&0.1&1.5733&3&1.3327&2.9&1.0942&24.7333&1.0996&9.1333&1.0995&25.9333\\
&0.2&1.6912&6.3333&1.3843&4.4333&1.1922&39.4333&1.1994&13.1&1.1992&26\\
\midrule \multirow{3}{*}{ch150}&0.05&1.6411&1.9667&1.494&1.8&1.0467&8.7&1.0497&11.2&1.0497&21.8333\\
&0.1&1.6289&3.2&1.5463&3.3667&1.0963&30.0333&1.0997&11.9&1.0996&36.7\\
&0.2&1.728&7.4667&1.613&6.4&1.1959&47.7667&1.1994&11.9667&1.1994&37\\
\midrule \multirow{3}{*}{tsp225}&0.05&1.6566&2.5333&1.717&2.1667&1.048&8.1&1.0498&9.8&1.0498&10.5\\
&0.1&1.7028&4.1&1.775&4.6&1.0978&36&1.0997&12&1.0997&54.8\\
&0.2&1.7955&7.9667&1.834&8.1667&1.1974&70.8&1.1994&17.3&1.1995&56\\
\midrule \multirow{3}{*}{pcb442}&0.05&1.8287&2.3333&2.3609&4.1333&1.0485&9.2&1.0498&12.4&1.0498&15.2667\\
&0.1&1.8671&4.0333&2.3791&7.4333&1.0985&43.1333&1.0998&17.2333&1.0998&109.6333\\
&0.2&1.9747&8.0667&2.4711&12.7&1.1983&162.8667&1.1996&17.5333&1.1997&110\\
\bottomrule
\end{tabular}
\end{table*}

\section{Conclusions}\label{sec:con}

In this work, we explored an application of a niching-based method in finding approximations to TSP instances that maximize diversity. As TSP is hard to approximate, we used a 2-stage approach, where the first stage uses a simple niching memetic algorithm to discover distant satisfactory solutions, and the second stage focuses on diversification with $(\mu+1)$-EA. Our NMA is a derivation of the state-of-the-art for multi-solution in TSP \cite{Huang2019}, with simplification and a randomized improvement-first variant of local search instead of greedy. Our experiment on TSPLIB instances showed that our approach improves the sum-sum diversity consistently compared to pure $(\mu+1)$-EA with optimal seeds, but falls short in the sum-min diversity in some cases due to clustering arising from niching. We also found that our NMA is significantly superior than the original in terms of the number of found distant solutions, quality of found solutions, and running time. The observations indicate the key role of local search in the performance of NMA, since such improvements came mostly from modifying the local search heuristic to be more suitable to the problem.

Due to the approach's simplicity, there is still much room for improvements. As seen on larger instances, our NMA exhibits scaling issues, which may be addressed by more powerful crossover, mutation and local search heuristics. This could allow to maintain low convergence time with larger populations, which might lead to discovering more feasible local optima regions, and higher diversity. Furthermore, there are many other niching techniques, which may be tailored in ways that eliminate the need for a separate diversification stage.

% \begin{acks}
% Some acknowledgements ...
% \end{acks}
\section*{Acknowledgment}
This work has been supported by the Australian Research Council (ARC) through grants DP190103894, FT200100536, and by the South Australian Government through the Research Consortium ``Unlocking Complex Resources through Lean Processing''.

\bibliographystyle{unsrt}
\bibliography{refs}
%%%%%%%%%%%%%%%%%%%%%%%%%%%%%%%%%%%%%%%%%%%%%%%%%%%%%%%%

%%%%%%%%%%%%%%%%%%%%%%%%%%%%%%%%%%%%%%%%%%%%%%%%%%%%%%%%

\end{document}